\documentclass[letterpaper, 10 pt, conference]{./ieeeconf}

\IEEEoverridecommandlockouts                              

\usepackage[utf8]{inputenc}
\usepackage{graphicx}
\usepackage{url}
\usepackage[linesnumbered]{algorithm2e}
\usepackage{subfig}
\usepackage{mathtools}
\usepackage{hyperref}
\usepackage{textcomp}
\usepackage[capitalise]{cleveref}
\usepackage{array}
\usepackage{booktabs}
\usepackage{makecell}
\usepackage[font=small]{caption}

\usepackage{gnuplot-lua-tikz}

\newcommand{\labelsubseccounter}[1]{
}
\usepackage[style=ieee,autocite=footnote,maxcitenames=1,minnames=1,maxalphanames=4,minalphanames=3,natbib=true,url=false, doi=true]{biblatex}

\DefineBibliographyStrings{english}{%
    andothers = {et\addabbrvspace al\adddot}
}
\begin{document}

\setlength{\textfloatsep}{0pt}

\title{A method to segment maps from different modalities using free space layout\\\textsc{maoris}: \underline{ma}p \underline{o}f \underline{ri}pples \underline{s}egmentation}

\author{Malcolm Mielle, Martin Magnusson, Achim J\@. 
Lilienthal
   \thanks{Center of Applied Autonomous Sensor Systems (AASS), 
{\"O}rebro University, Sweden.
     {\tt firstname.lastname@oru.se}}%
   \thanks{This work was funded in part by
     the EU H2020 project
     SmokeBot (ICT-23-2014 645101) and by the Swedish Knowledge Foundation under contract number 20140220 (AIR)
}}

\maketitle

\begin{abstract}

How to divide floor plans or navigation maps into semantic representations, such as rooms and corridors, is an important research question in fields such as human-robot interaction, place categorization, or semantic mapping. While most works focus on segmenting robot built maps, those are not the only types of map a robot, or its user, can use. We present a method for segmenting maps from different modalities, focusing on robot built maps and hand-drawn sketch maps, and show better results than state of the art for both types.


Our method segments the map by doing a convolution between the distance image of the map and a circular kernel, and grouping pixels of the same value. Segmentation is done by detecting ripple-like patterns where pixel values vary quickly, and merging neighboring regions with similar values.

We identify a flaw in the segmentation evaluation metric used in recent works and propose a metric based on Matthews correlation coefficient (MCC). We compare our results to ground-truth segmentations of maps from a publicly available dataset, on which we obtain a better MCC than the state of the art with 0.98 compared to 0.65 for a recent Voronoi-based segmentation method and 0.70 for the DuDe segmentation method.
%
We also provide a dataset of sketches of an indoor environment, with two possible sets of ground truth segmentations, on which our method obtains an MCC of 0.56 against 0.28 for the Voronoi-based segmentation method and 0.30 for DuDe.

\end{abstract}


%
\IEEEpeerreviewmaketitle

\section{Introduction}


State-of-the-art SLAM algorithms enable robots to build metric maps representing their environment. However, robot built maps (referred to as robot maps) are not always easy to understand. They suffer from sensor noise, clutter, and can be complex if the environment is large and complicated. It is often easier for a human to use high-level features, such as rooms and corridors, than to directly be presented with the robot map. This is true in fields such as human-robot interactions, where one could communicate places and directions using rooms' names instead of coordinates, or in planning, where robots need  to visit rooms in the optimal order.



But robot maps are not the only type of map a robot can use. 
%
%
%
For example, hand-drawn sketches are very intuitive interfaces for human-robot interactions and are effective at conveying spacial configurations, as shown by \textcite{skubic_using_2007}. However, sketch maps are not metrically accurate, sometimes by design. For example, the person who drew might have altogether ignored a feature of the environment because they judged it unimportant.
While humans are able to take those changes into account, and understand the abstraction in the drawing, a robot can not easily interpret such a fuzzy representation of an environment. 

\begin{figure}[t]
    \begin{center}
        \includegraphics[width=0.35\textwidth]{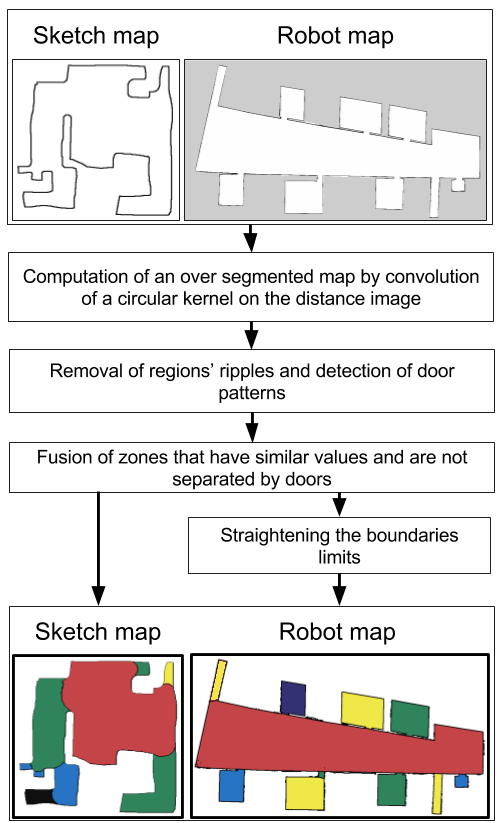}
        \caption{\textsc{maoris} map segmentation process illustrated for a sketch map and a robot map of 2 different environments.}
        \label{fig:process}
    \end{center}
\end{figure}

In our work, we present a novel method for segmenting maps, even when coming from very different modalities, and, as an application, we focus on robot and sketch maps. An outline of the method is shown in \cref{fig:process}.
The algorithm was evaluated on datasets representing both robot maps and sketch maps, and compared to ground truth segmentations given by users.
The contributions of this paper are:
\begin{itemize}
    \item A novel method to extract regions from maps from different modalities.
    Our method is more robust than the state of the art.
    \item A dataset of sketches representing three indoor places with human labeled segmentations.
    \item A discussion on how to evaluate and compare the results of map segmentation algorithms, and a proposed new metric for doing so.
\end{itemize} 
\section{Related Work}

\textcite{bormann_room_2016} review the literature on room segmentation, and select and implement four algorithms as ROS packages. They provide and compare those methods against 20 floor plans; in most cases a segmentation based on the Voronoi diagram gives the best approximation of the ground truth segmentation. However, the Voronoi based segmentation tends to over-segment the regions, especially corridors. Furthermore, the noisy nature of sketch maps makes it hard to extract the major skeleton of the Voronoi, resulting in over or under-segmentation. 

\textcite{fermin-leon_incremental_2017} propose the DuDe segmentation method that uses a contour based part segmentation~\cite{liu2014dual} for the construction of 2D topological maps. They develop a batch and an incremental algorithm for segmentation and tested them against the ones presented by \textcite{bormann_room_2016}. 
While Voronoi-based segmentation performs better, their method is faster. However, their method over-segments corridors. Also, it is based on contour shape and, when used on sketches with less attention to details, it leads to poor segmentation results as shown in \cref{subsec:sketch}.

\textcite{fabrizi_extracting_2000} use image processing to extract features from an occupancy grid constructed from a sonar. They use fuzzy opening and closure to compute information about the free space and use a watershed algorithm to create regions. Regions are classified into rooms and corridors according to their eccentricity. The watershed uses door patterns to avoid fusing different rooms together. However, those door patterns are not always present in maps.

\textcite{park_map_2016} extract maximal empty rectangles in maps and use them for map matching. Since orientation, accuracy, and scale of the maps are unknowns, the final merged map is estimated by minimizing the differences of the factors. 
This method is dependent on the presence of orthogonal angles in the map, which is not realistic for sketch maps.

\textcite{ahmed_automatic_2012, de_las_heras_statistical_2014} discuss a system that uses architectural floor plans with symbols and textual annotations. 
Their method is made for floor plans with high accuracy and a lot of details, e.g. different types of walls and labels, and is not applicable for sketch and robot maps. 

\textcite{diosi_interactive_2005} implement a semi-autonomous room segmentation algorithm. A robot maps an environment while following a user who gives it labels for different locations. Once the mapping is completed, the distance image is used to generate local maxima and they use gradient ascent to group all pixels that move to the same local maxima into regions. The unlabeled segments are merged into the labelled segment that minimizes the distance from the centroid of the unlabelled segment to the closest label. Since this method depends on previous labeling by a user and the position of labels on the map, it is thus not adaptable for sketches and general use cases.


\section{Map segmentation}

\begin{algorithm}[t]\small
  \vspace{1mm}
	\KwData{\textit{map}}
	\KwResult{\textit{segmented map}}
	
	Calculate distance image from map\;
	Calculate free space image\;
	Group adjacent pixels of same value in regions\;
	Remove ripples\;
	Merge regions with similar values and not separated a door\;
	Remove regions created by thick walls\;
	\If{map is a robot map}{
	    Straighten boundaries\;
	}
	
	\Return Segmented map\;
	 	
	\caption{Segmentation algorithm.}
	\label{algo:segmentation}
	\vspace{1mm}
\end{algorithm}

%

We hypothesize that regions in indoor maps can be found by looking at the layout of free space; more specifically, by looking at changes of sizes between rooms,
e.g doors in between rooms, or rooms and corridors with different sizes. Previous work like the distance transform-based segmentation of~\textcite{bormann_room_2016} used a threshold on the distance image to extract the maximum number of regions. Thus, finding a good value for the threshold depends on the size difference between the biggest and the smallest regions. Furthermore, that threshold is arbitrarily chosen by a user depending on the environment. 
On the other hand, our method uses the distance image to produce a new image where a pixel's value represents the size of the region it belongs to. By looking at this image, our method is independent of the shapes of regions, and only depends on the size differences \emph{between adjacent regions}, which is a more relevant measure than the difference between the biggest and smallest regions.

In this section, we present the algorithm we use to segment maps, which is described in \cref{algo:segmentation}. 
The key elements explained in the remainder of this section, are:
A) extraction of an over-segmented map from the layout of the free space,
B) removal of a particular type of region that we call ripples,
C) further refining of the segmentation by merging neighbors with similar value and detecting door patterns, if any are present,
D and~E) removal of regions created by thick walls and contour straightening for robot maps.

\subsection{Computation of the free space image}
\label{subsec:freespace}
\labelsubseccounter{subsec:freespacesub}

We first compute an image where each pixel value represents the size of the region it belongs to, referred to as the free space image (FSI).
Effectively, the FSI represents the amount of free space surrounding every pixel in the image. 

The algorithm to extract the FSI is shown in \cref{algo:freespace}. It starts with calculating the distance image of the map, i.e. the image where each pixel's value is the distance to its closest obstacle.
For each pixel of the distance image, we create a circular mask centered on the pixel and whose radius is the value of the pixel. For every pixel in the circular mask, if the value of the equivalent pixel in the FSI is less than the circle radius, the FSI pixel's value is changed to the radius of the circle. Once all pixels have been considered, neighboring FSI pixels with the same value are grouped into regions.
%

Since we do not assume any particular direction for regions in the map we used a circular mask, but other shapes can be used for other application, e.g. a squared shape for environments with only orthogonal angles. An example of sketch map's FSI can be seen in \cref{fig:freespace}.

\subsection{Merging of ripple regions}
\labelsubseccounter{subsec:ripples}

\begin{figure}[t]
\vspace{-3mm}
\centering
\subfloat[Original sketch map.]{\includegraphics[width = 0.3\columnwidth]{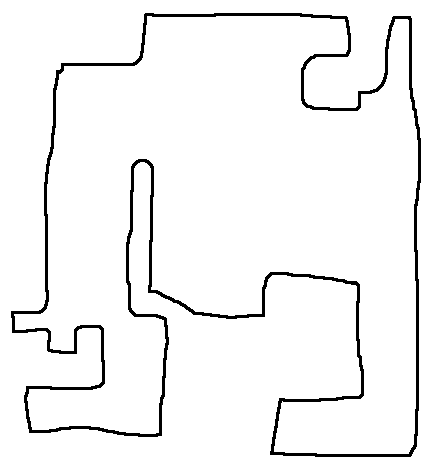}\label{fig:model}}\,
\subfloat[Free space map. The darker the pixel, the higher its value.]{\includegraphics[width = 0.3\columnwidth]{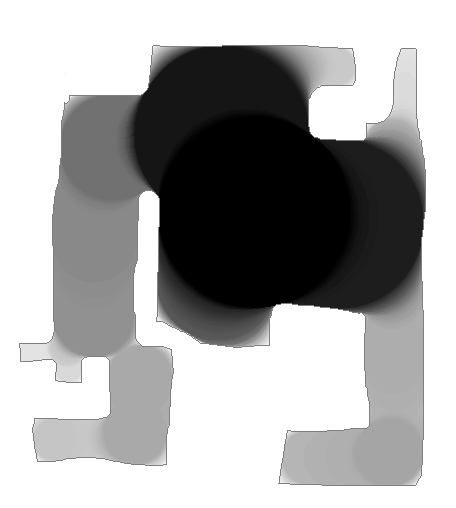}\label{fig:freespace}}\,
\subfloat[Neighboring pixels of same value are grouped together in regions.]{\includegraphics[width = 0.3\columnwidth]{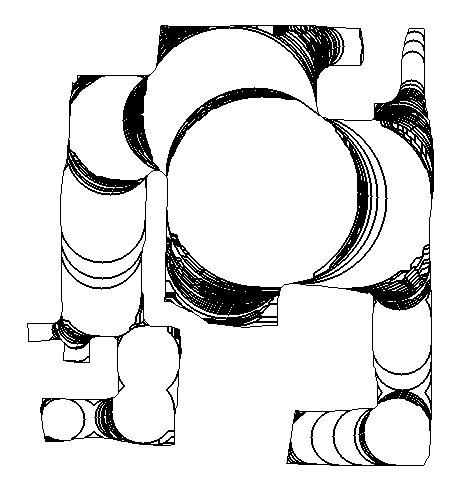}\label{fig:freespaceregion}}\\
\subfloat[All ripples are removed from \cref{fig:freespaceregion}.]{\includegraphics[width = 0.3\columnwidth]{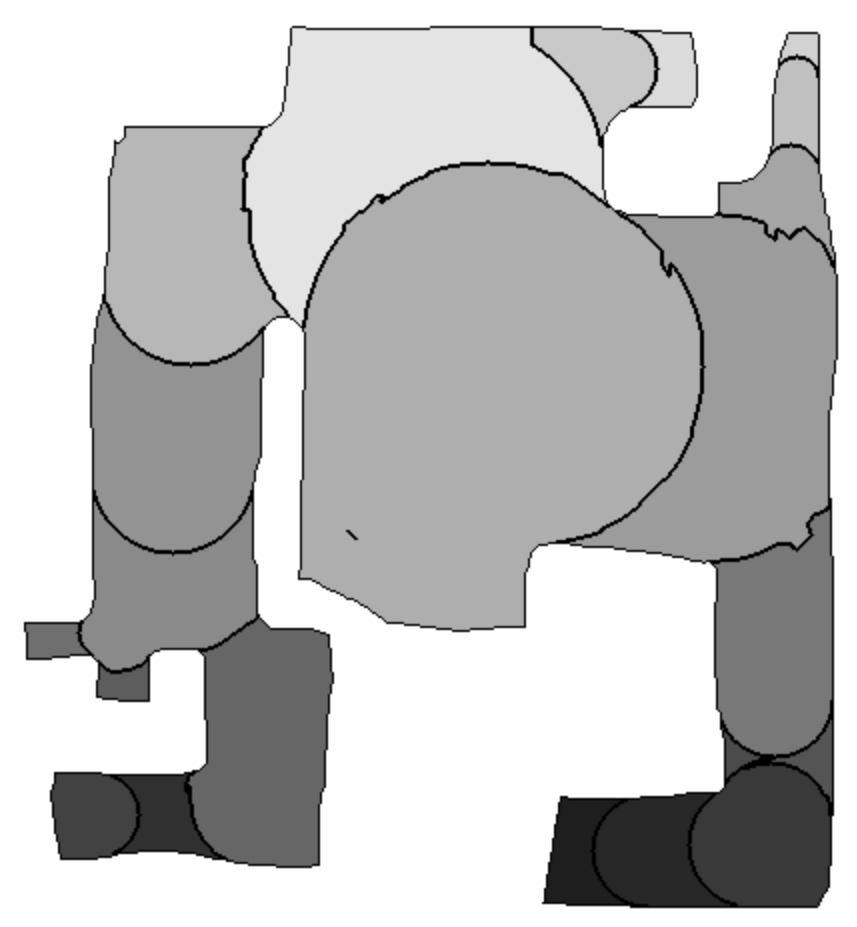}\label{fig:ripples}}\,
\subfloat[Regions found from \cref{fig:ripples} by merging regions with similar values and no door between them.]{\includegraphics[width = 0.3\columnwidth]{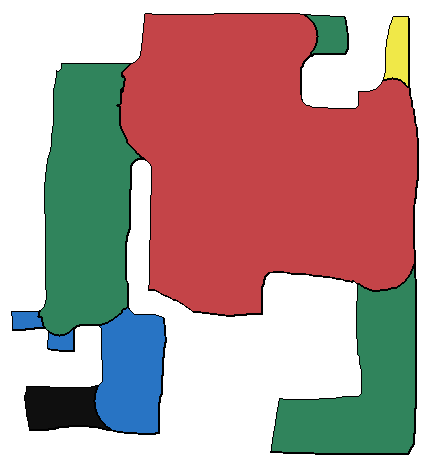}\label{fig:watershed}}\,
\subfloat[Final regions after boundaries straightening.]{\includegraphics[width = 0.3\columnwidth]{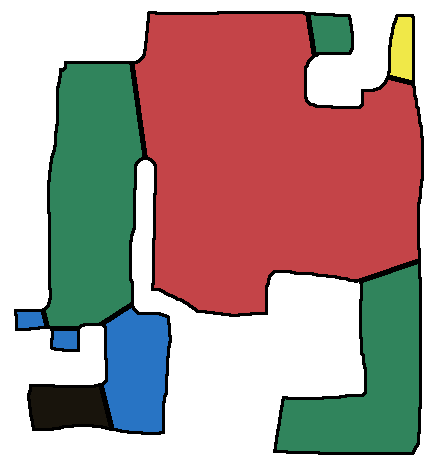}\label{fig:finalimg}}\,
\caption{All steps from the segmentation algorithm.}
\label{}
\end{figure}

The FSI provides an over-segmented map where variations of size, however small, are represented (see \cref{fig:freespaceregion}). Small variations in pixel value create ripples, i.e very thin regions that vary slightly in pixel values.
Those are often present between bigger regions and the second step of the algorithm removes them by merging them into larger regions.


However, it is important that ripples are merged into the region they belong to. Otherwise, two ripples could be merged together, creating a non-ripple region where there shouldn't be any, over-segmenting the map. Since ripples are smaller than the region they belong to, by considering regions from the one with the highest pixel value to the lowest, the algorithm merges large ripples into large regions first, before merging the smaller ripples into the remaining regions, thus avoiding over-segmentation.


Only a simple rule is needed to detect ripples. Since ripples are elongated, around half of their contour is in contact with an adjacent region, i.e a neighbor region. If more than 40\% of the contour of a region is in contact with a neighbor region, the region is merged with its neighbor. If a region is a ripple of more than one region, it is merged with the region that has the closest pixel value.
%
Every time a ripple is merged, all neighboring regions of the ripple are checked again to see if they became ripples of the new, merged, region. A stability analysis is shown in \cref{subsec:stability} confirming that using 40\% leads to good segmentations of the maps.
%

 
%
The segmentation of a sketch map, after ripple merging, can be seen in \cref{fig:ripples}.

\begin{algorithm}[t]\small
  \vspace{1mm}
	\KwData{\textit{map image}}
	\KwResult{\textit{FSI}}
	
	dimage = distance image from map\;
	fsi = empty image\;
	\For{each pixel in dimage}{
	    create circle with radius pixel value\;
	    \For{each pixel in circle}{
	        \If{the value of the equivalent pixel in fsi $\leq$ circle radius}{
	            equivalent pixel in fsi = circle radius\;
	        }
	    }
	}
	
	\Return fsi\;
	 	
	\caption{Free space image algorithm}
	\label{algo:freespace}
	\vspace{1mm}
\end{algorithm}

%
%

\subsection{Merging of neighbor regions with similar pixel values}
\labelsubseccounter{subsec:watershed}


At this point, the map segmentation still has regions that belong to the same place, that are not ripples,
but need to be merged together, e.g. corridors that slowly become smaller will be cut into multiple parts of different sizes. This is visible in \cref{fig:ripples} where the left corridor is still separated in multiple regions because its width is not constant. Thus, this step of the algorithm merges neighboring regions with similar pixel values.

We define $pV1$ and $pV2$ the pixel values of two regions, $t\_merging$ the merging threshold, and $m$ a margin added to the merge threshold. $t\_mergin$ is a number between 0 and 1 representing the relative pixel-value difference between regions for automatic merging, with 0 being no merging possible and 1 being always merge regions. Hence, the merging threshold and margin 
represent how similar in value two regions need to be before being considered for merging.

%
We consider regions from the one that contains the most pixels to the one that contains the least, and recursively look at all neighbor regions. If the value difference $|pV1 - pV2|$ is less than the highest pixel value multiplied by the threshold, i.e. \cref{eq:mergecomp1} holds, both regions have similar pixel values and are merged. 
\begin{equation}  \label{eq:mergecomp1}
    |pV1 - pV2| \leq \max{(pV1, pV2)} * (t\_merging)
\end{equation}
However, using a fixed threshold will not merge all regions that should be merged, due to the fuzzy nature of regions. Thus, for two neighboring regions, if $|pV1 - pV2|$ is more than the highest pixel value multiplied by the threshold, i.e. \cref{eq:mergecomp1} does not hold, but it is less than the same equation with a margin $m$ added to the threshold, i.e. \cref{eq:mergecomp2} holds, the algorithm checks if the regions should be merged by studying their neighborhoods.
It will merge the regions if and only if one of the two regions has a similar pixel value to at least one neighbor of the other region, i.e \cref{eq:mergecomp1} holds between a region and at least one neighbor of the other region. 
\begin{equation}  \label{eq:mergecomp2}
    |pV1 - pV2| \leq \max{(pV1, pV2)} * (t\_merging + m)
\end{equation} 

For regions between which ripples were present, the algorithm does not consider them for merging if the ripples were created by a door. Doors can be found by looking at the minimum value of all the ripples that were present between two regions and making sure that the difference in pixel values between this minimum value and both regions was not significant, i.e. \cref{eq:mergecomp1} holds between the minimum and both regions.

Stability analyses for $t\_merging$ and $m$ can be found in \cref{subsec:stability}. The final segmented version of \cref{fig:model} can be seen in \cref{fig:watershed}

\subsection{Taking into account wall thickness}
\labelsubseccounter{subsec:wallthick}

%

Thick walls can over-segment the map by creating small regions, e.g. where there are doors. Instead of assuming that we know the walls' thickness, the algorithm considers that every region with more than a certain percentage $d\_threshold$ of its perimeter in contact with other regions belongs to another region.  Thus, those regions are fused into a neighbor region with less than $d\_threshold$ of its perimeter in contact with other regions.
A stability analysis of this parameter
can be found in \cref{subsec:stability}.


%

\subsection{Straightening the boundaries}
\labelsubseccounter{subsec:refining}

The contact line between regions depends on the shape of the mask used in \cref{subsec:freespace}. A circular shape works well with sketch maps since the drawing is inaccurate and boundaries between regions might not be straight. However, for metrically correct robot maps, boundaries between regions can be straightened to increase the accuracy of the segmentation.
We replace each boundary between regions by the straight line between its endpoints. 

While this refining step is not applied to sketch maps,
for illustration purposes, a straightened segmented version of \cref{fig:model} can be seen in \cref{fig:finalimg}.

\section{Experiments}
\label{sec:experiment}

\begin{figure}[t]
  \vspace{+2mm}
  \centering
  \begin{tabular}{  c  m{2cm}  m{2cm}  }
  \toprule

    Example segmentation & Our method & Bormann's method \\
    \cmidrule(lr){1-3}
    \begin{minipage}{.3\linewidth}
      \includegraphics[width=\linewidth]{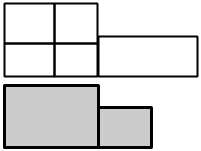}
    \end{minipage}
    &
    \makecell{
    $\frac{1+0.5+0+0+0}{5}$\\
    $= \frac{1.5}{5}$\\
    $= 0.3$
    }
    & 
    \makecell{   
    $\frac{1+1+1+1+0.5}{5}$\\
    $= \frac{4.5}{5}$\\
    $= 0.9$
    }
    \\
    \bottomrule
  \end{tabular}
  \caption{Difference of recall between our method and Bormann's method. The top image is the ground truth while the bottom one is the under-segmented map. With Bormann's calculation, all subregions in the biggest room increase the mean recall by being considered perfectly fitted in the biggest segmented region: four perfect recalls for one segmented region. The under-segmented room weights more on the final recall result than the corridor. With our measure, only one subregion is associated with the biggest room and all other subregions yield a score of 0, driving the recall down.}\label{tbl:recall}
    \vspace{+1mm}
\end{figure}


We compare our method (public implementation is available online\footnote{\url{https://github.com/MalcolmMielle/maoris}}) with approaches presented by \textcite{bormann_room_2016} and against the DuDe method of \textcite{fermin-leon_incremental_2017}, using the dataset of \textcite{bormann_room_2016} and against our own dataset of sketches~\cite{mielle_malcolm_2017_892062}. The dataset of \textcite{bormann_room_2016} is made of 20 maps, either without clutter or with artificially added clutter. Since our method is only adapted to environments without furniture, we used the non-cluttered images. Clutter can be removed in a preprocessing step or by extracting the walls of an environment, as done by \textcite{wulf20042d}; we are leaving those ideas for future work. 

\begin{figure}[t]
\vspace{+1.5mm}
    \begin{center}
        \includegraphics[width=0.15\textwidth]{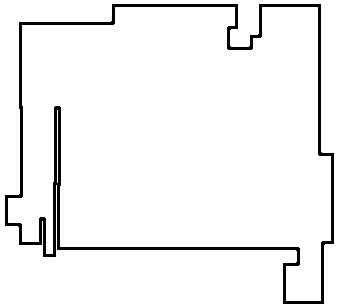}
        \caption{Model environments of KTH SLAM dataset explored and drawn by the users to produce the sketches in the dataset.}
        \label{fig:kth}
    \end{center}
\end{figure}

We make available a dataset of 25
sketches. Those sketches were obtained from a virtual environment in a web browser\footnote{\url{http://aass.oru.se/Research/mro/smokebot/sketchmap-web/}} and correspond to the ground truth of the KTH dataset for SLAM\footnote{\url{http://www.nada.kth.se/\%7Ejohnf/kthdata/dataset.html}} (\cref{fig:kth}).
Each sketch map is associated with two possible ground-truth segmentations, obtained by asking two non-expert users to segment the sketches. The only instruction was to segment regions so that, if a robot or person had to visit the environment, they would have seen every part of it by visiting every region of the map.

\textcite{bormann_room_2016} introduced a quality measure based on precision and recall, where precision of a segmented region is the maximum overlapping area of the segmented region with a ground truth region divided by the area of the segmented room, while recall of a ground truth region is the maximum overlapping area of the ground truth region with a segmented room divided by the area of the ground truth region. The full segmentation's precision is the mean of all segmented regions' precision while the recall is the mean of all ground truth regions' recall. A segmentation is good if both the precision and recall are high.

While this metric evaluates how the segmentation fits the ground truth, it is biased toward giving high recall results on under-segmentation, and high precision results on over-segmentation, as illustrated for recall in \cref{tbl:recall}. 
Indeed, the segmentation recall being the mean of the recall of all ground truth regions, one segmented region including multiple ground truth regions will have a lot of weight on the segmentation's recall. In some cases, it can virtually erase the influence of segmented regions as in \cref{tbl:recall}, where the central room recalls outweigh the corridors recall. A similar case can be made for the precision.
%

Furthermore, since the true positive values (the pixels that fit both in the ground truth and segmented regions) are not the same for the calculation of the precision and the recall, those measures make us unable to represent a confusion matrix or calculate meaningful metrics such as the F-score, G-score or Matthews correlation coefficient.

\begin{figure}[t]
  \vspace{1mm}
    \begin{center} 
      \input{param_eval_ripples}
      \vspace{-2mm}
      \caption{Evaluation of the sensitivity of the algorithm to the parameters for fusing ripples with $t\_merging$ = $0.30$ and $m$ = $0.1$. Points are the median MCC, while the whiskers represent the minimum and maximum value. When considering both map types, the best result is found when regions with 40\% of their contour in contact with another region are considered ripples.}
        \label{fig:ripples_p}
    \end{center}
    \vspace{-8mm}
\end{figure}

\begin{figure}[t]
\vspace{+2mm}
    \begin{center} 
      \input{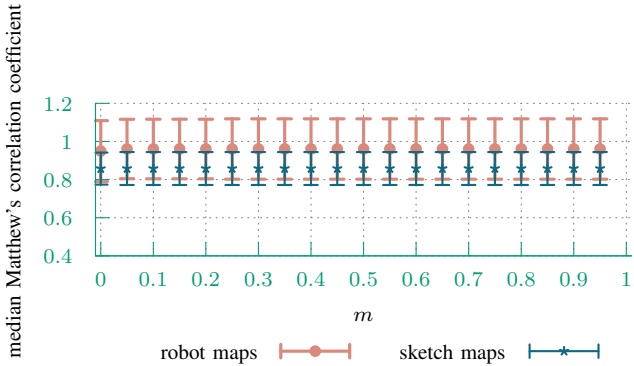}
      \vspace{-2mm}
      \caption{Evaluation of the margin $m$ with $t\_merging$ = $0.30$. One can observe that the algorithm is largely insensitive to the value of m, but having it different than 0 results in slightly better segmentation. We obtained similar results for other values of $t\_merging$.}
        \label{fig:meval}
    \end{center}
\end{figure}

\begin{figure}[t]
  \vspace{1mm}
    \begin{center} 
      \input{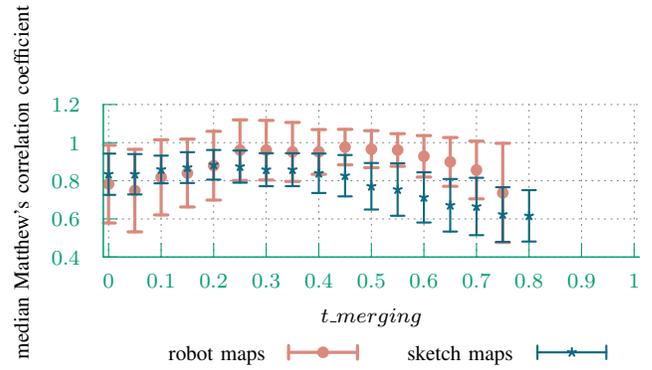}
      \vspace{-2mm}
      \caption{Evaluation of $t\_merging$ with $m = 0.1$. The best results are found between 0 and 0.3 for sketch maps and between 0.1 and 0.5 for robot maps. Thus a good value of $t\_merging$ for both would be 0.3.
      }
        \label{fig:teval}
    \end{center}
    \vspace{-2mm}
\end{figure}

\begin{figure}[t]
\vspace{+2mm}
    \begin{center} 
      \input{param_eval_doors}
      \vspace{-2mm}
      \caption{Evaluation of the sensitivity of the algorithm to the parameters for fusing regions created by thick walls with $t\_merging$ = $0.30$ and $m$ = $0.1$. Sketch maps' segmentations' results do not depend much on this parameter: they are good above 40\%, with the highest score at 60\%, and a very small decrease until 100\% where $d\_threshold$ is not used. However, it is an important factor for robot maps: the best results are found between 15 and 60\% with a sharp decrease of the median Matthews correlation coefficient after.}
        \label{fig:doors}
    \end{center}
\end{figure}

We introduce another way to evaluate the segmentation results borrowed from clustering measures. Each segmented region is associated with the ground truth region with the biggest overlap and that is not already associated with a segmented region. The true positives (tp) are all pixels in both the segmented region and the ground truth, the false positives (fp) are the pixels in the segmented region but not in the ground truth, the false negatives (fn) are all the pixels in the ground truth region but not in the segmented, and the true negatives (tn) are all the pixels in neither the ground truth nor the segmented region. To be able to compare the different results given by each segmentation algorithm, we use the Matthews correlation coefficient (MCC):
\begin{equation}  \label{eq:matthews}
    MCC = \frac{tp * tn - fp *fn}{\sqrt{(tp+fp)(tp+fn)(tn+fp)(tn+fn)}}
\end{equation} 

The MCC ranges between $-1$ and $1$, with the best predictive result being $1$, $0$ being no better than guessing, and $-1$ indicating total disagreement. One advantage of the MCC for evaluating map segmentation is that it stays balanced even when the classes are of different sizes.
While there is no best way to represent a confusion matrix with a single number, the MCC is generally regarded as one of the best measures to do so~\cite{powers_evaluation:_2011}. Regions that were not associated with a corresponding region have a MCC of 0. The final MCC of a segmented map is the mean of all regions' MCC. A public implementation of the evaluation program can be found online with the \textsc{maoris} package.



%


\subsection{Evaluation of the parameters' stabilities}
\label{subsec:stability}

We did a stability analysis of every parameter of our method using median MCC as a measure of segmentation goodness. We ran the analyses over the 16 smallest uncluttered maps of Bormann's dataset and all maps in our sketch dataset against one of the user's ground truth segmentation. 

We first confirmed that using 40\%, as the threshold value above which a region is determined to be a ripple, was a valid assumption. As can be seen in \cref{fig:ripples_p}, the MCC is best between 30\% and 45\%, with the highest value at 40\%, confirming the validity of the threshold's value.

\begin{figure}[t]
\vspace{+2mm}
    \begin{center} 
      \input{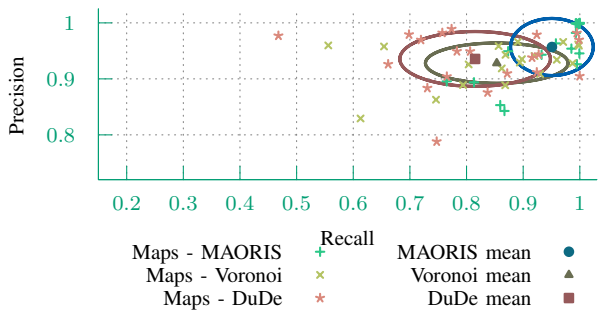}
      \vspace{-2mm}
      \caption{Precision and recall for every unfurnished map in \textcite{bormann_room_2016} dataset. One can see that \textsc{maoris} outperforms the Voronoi segmentation and DuDe both in precision and recall. \textsc{maoris} has a median MCC of 0.98, while Voronoi has 0.65 and DuDe has 0.70. 
}
        \label{fig:mapresult}
    \end{center}
\end{figure}

\begin{table}[t]\normalsize
  \centering
  \begin{tabular}{ c c c c }
  \toprule
    
   Dataset & \textsc{maoris} & Voronoi & DuDe \\ 
   \cmidrule(lr){2-4}
   Bormann's & 0.98 & 0.65 & 0.70 \\
   Sketch & 0.56 & 0.28 & 0.30 \\ 
   \bottomrule
  \end{tabular}
  \caption{Median Matthews correlation coefficient for each method on all maps of both datasets. The closer this coefficient is to 1 the better the segmentation is.}\label{tbl:mcc}
  \vspace{+2mm}
\end{table}

We then evaluated the segmentation results' stability to $t\_merging$, the threshold for fusing adjacent regions with similar pixel values, and the margin $m$. As one can see in \cref{fig:meval}, the algorithm is not sensitive to the value of the margin $m$ for it has no influence on the sketch maps segmentation and only slightly increases the robot maps' segmentation's MCC when different than 0.
In our work, we chose to have $m = 0.1$.
The analysis for the threshold $t\_merging$ can be seen in \cref{fig:teval}. One can see that the segmentation goodness increases with $t\_merging$ but decreases passing a certain point: around $t\_merging$ = $0.5$ for the robot maps and $t\_merging$ = $0.3$ for the sketch maps. Also, one should note that the MCCs over the robot maps is constant between $t\_merging$ = $0.2$ and $0.4$. Those different values between sketch and robot maps come from the presence of doors in between regions in robot maps. Doors create a separation between regions, even when they are of similar size, allowing the merging threshold to be higher without merging regions that shouldn't be merged.
However, sketch maps in our dataset have no doors, leading to lower median MCC values above $t\_merging$ = $0.3$.
To get good results
for both sketch and robot maps, we chose $t\_merging$ = $0.3$. Hence, regions between which the difference of size is more than 1/3 of the size of the biggest region are not merged together.

Finally, we evaluated the influence of \emph{d\_threshold} on the segmentation results. One can see in \cref{fig:doors} that sketch map segmentations are not sensible to this parameter: the median MCC of the sketch maps segmentations is good above 40\% and its value increases until 60\% after which it only slightly decreases until 100\%. We hypothesize that users took into account the size of the pen while drawing, and thus, there are no regions created by the wall thickness.
On the other hand, \emph{d\_threshold} is an important parameter for robot maps:
the best results are obtained between 30\% and 60\%, and stays correct until 100\%.
For applications where one doesn't know if the wall thickness needs to be taken into account or not, we recommend 40\% as the best guess. In our work, we chose $d\_threshold$ to be 40\% for robot maps and 60\% for sketch maps.

\subsection{Segmentation results on Bormann's dataset}

\begin{figure}[t]
\vspace{+2mm}
    \begin{center} 
      \input{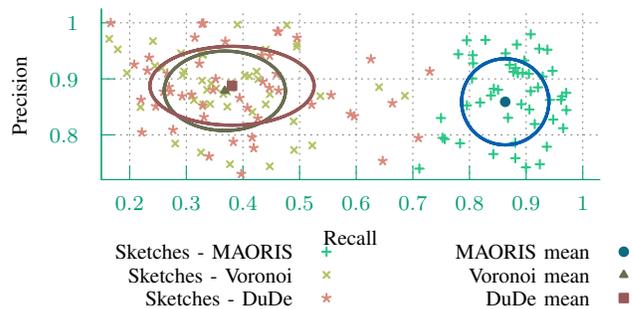}
      \vspace{-2mm}
      \caption{Precision and recall for each sketch in the dataset using batch DuDe, Voronoi segmentation, and our method for both evaluated using two ground truths. Ellipses represent the standard deviation on the recall and precision for each distribution. Our method outperforms the other methods with a median MCC of 0.56, while Voronoi has 0.28 and DuDe has 0.30.}
        \label{fig:sketches}
    \end{center}
\end{figure}

We ran our algorihtm (\textsc{maoris}) on the 20 uncluttered maps proposed by \textcite{bormann_room_2016}.
%
The result in precision and recall for each map can be seen in \cref{fig:mapresult} and MCCs are shown in \cref{tbl:mcc}. The Voronoi based segmentation of \textcite{bormann_room_2016} obtained an MCC of 0.65, 
 and the DuDe method of \textcite{fermin-leon_incremental_2017} obtained an MCC of 0.70.
In comparison, our method got a median MCC of 0.98, 
thus performing better than the two other methods. 

Examples of segmentations given by all three algorithms on maps from Bormann's dataset can be seen in \cref{fig:segex}.


\subsection{Segmentation results on our dataset of sketch maps}
\label{subsec:sketch}
We ran the three algorithms on 25 sketch maps and two user-provided ground truths. 
\textsc{maoris} performs better at segmenting sketch maps, with an MCC of 0.56, compared to the Voronoi segmentation with an MCC of 0.28 and the DuDe method with an MCC of 0.30.
When looking at \cref{fig:sketches}, one can see that both the Voronoi segmentation and DuDe have high precision and low recall. On the other hand, \textsc{maoris} has a better balance of precision and recall. 

Taking into account that sketch maps in our dataset do not have doors, it should also be noted that, with an MCC of 0.56, \textsc{maoris} does not depend on doors between regions for segmentation, as opposed to previous work~\cite{fabrizi_extracting_2000}.

\begin{figure*}[t]
\vspace{-3mm}
\centering
\subfloat[\textsc{maoris}]{\includegraphics[width = 0.23 \columnwidth]{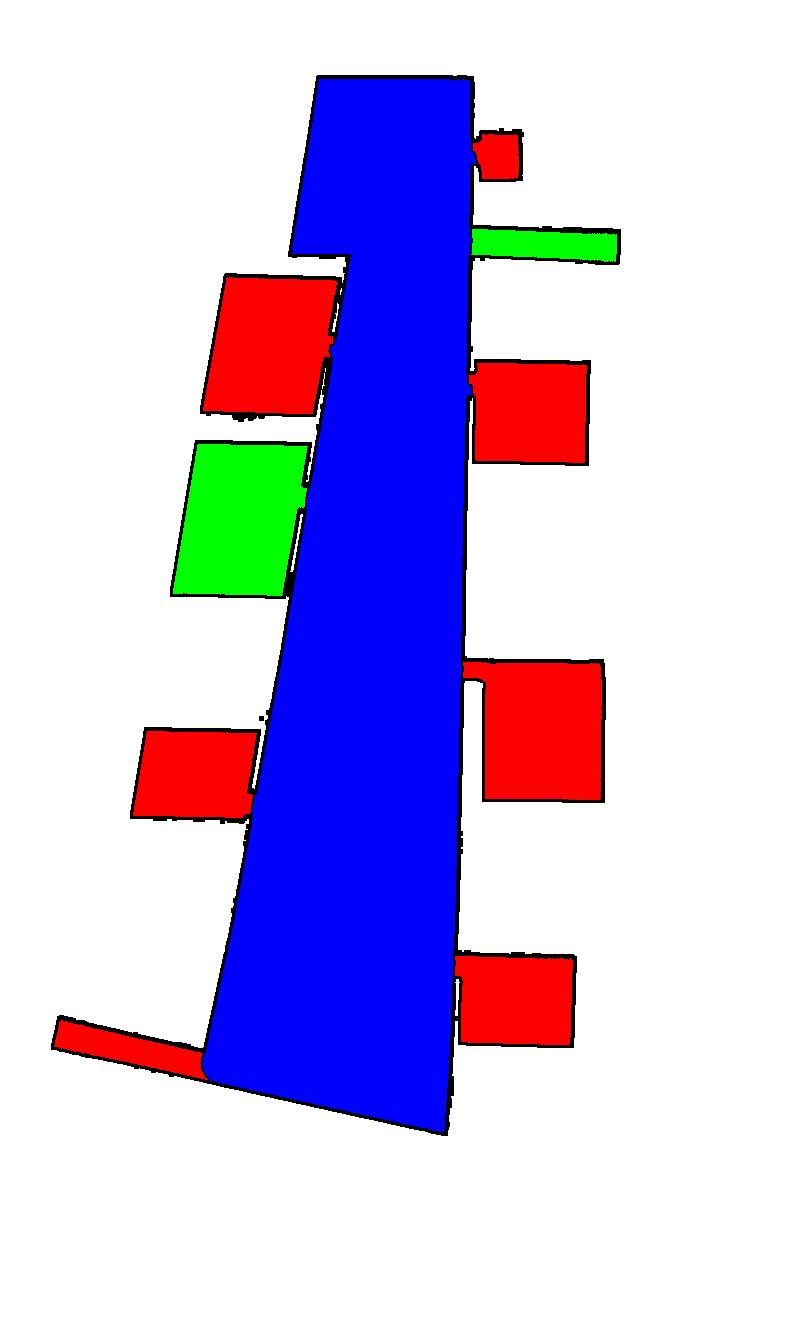}}\,
\subfloat[Voronoi]{\includegraphics[width = 0.23\columnwidth]{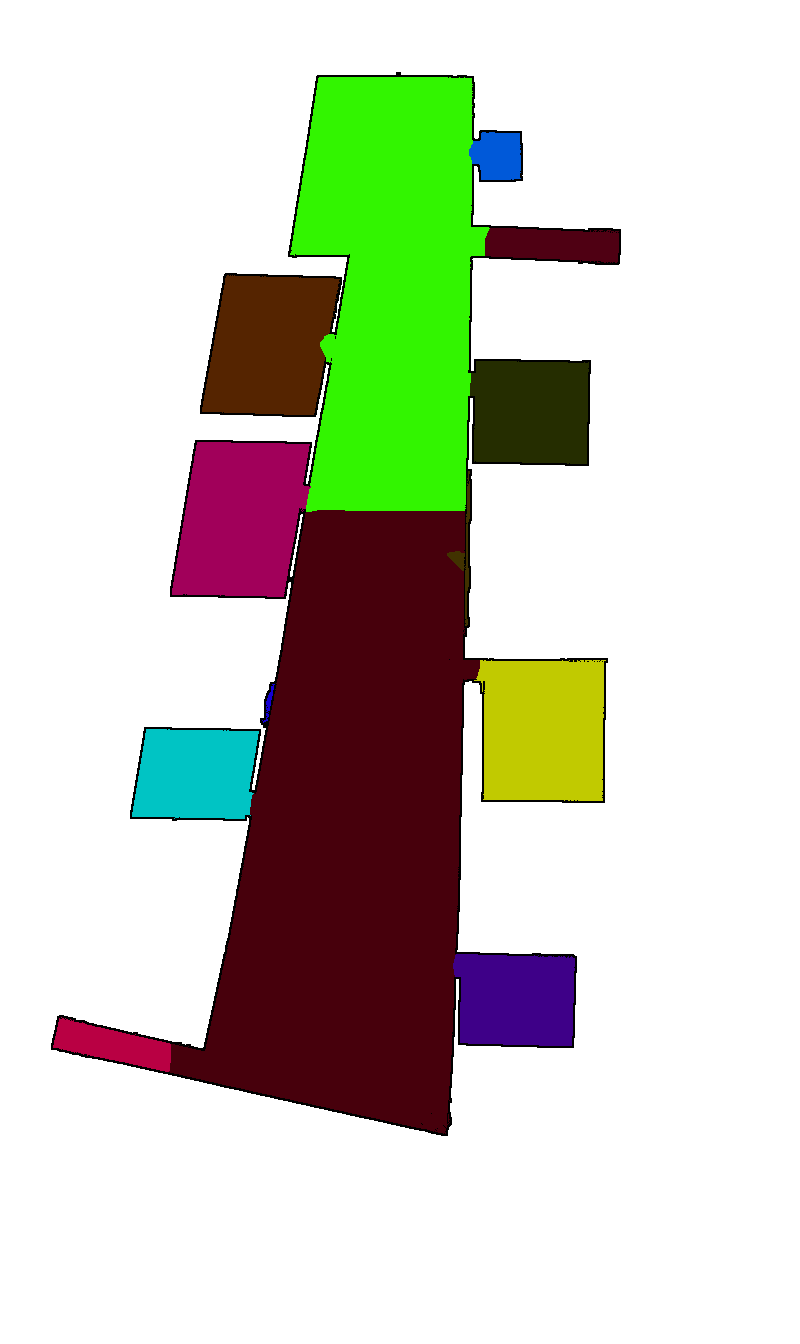}}\,
\subfloat[DuDe]{\includegraphics[width = 0.23\columnwidth]{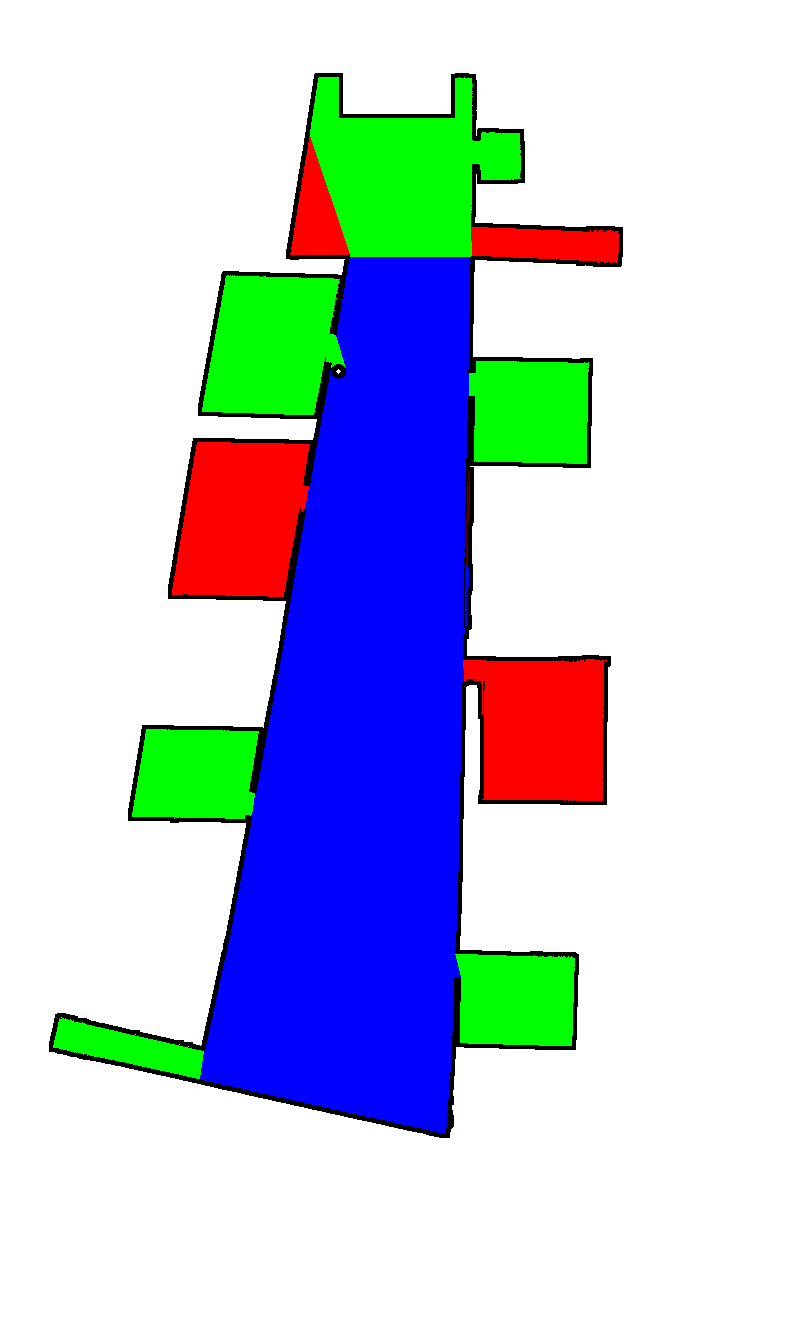}}\,
\subfloat[Ground truth]{\includegraphics[width = 0.23\columnwidth]{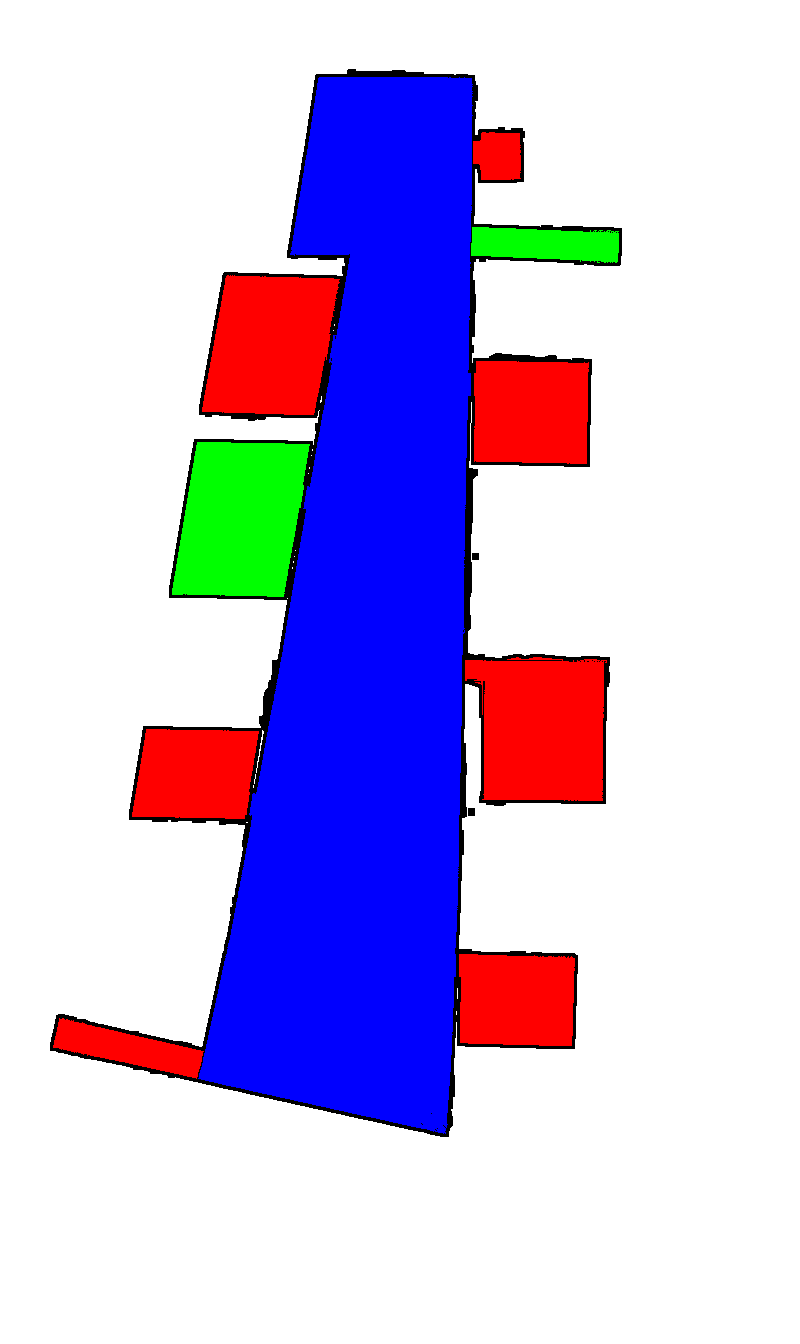}}\,
\subfloat[\textsc{maoris}]{\includegraphics[width = 0.23\columnwidth]{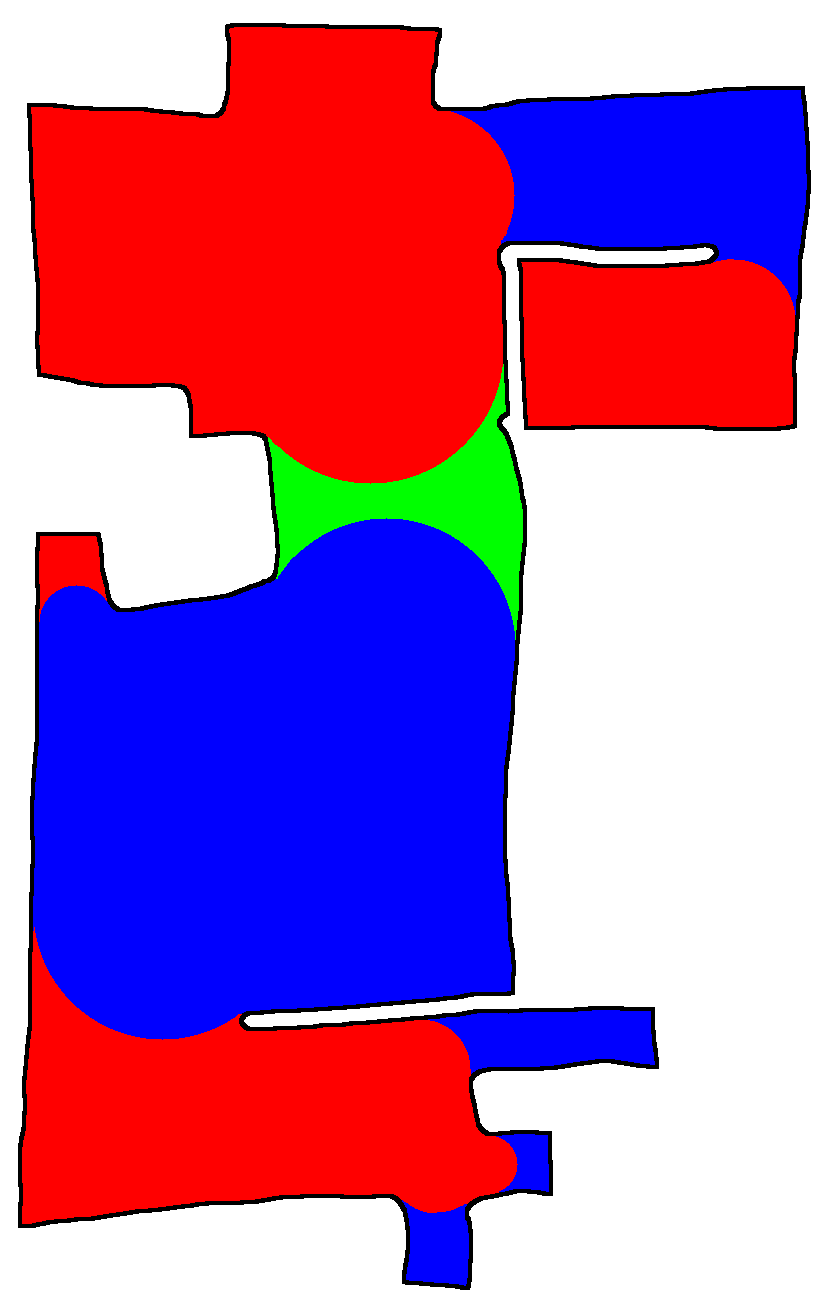}}\,
\subfloat[Voronoi]{\includegraphics[width = 0.23\columnwidth]{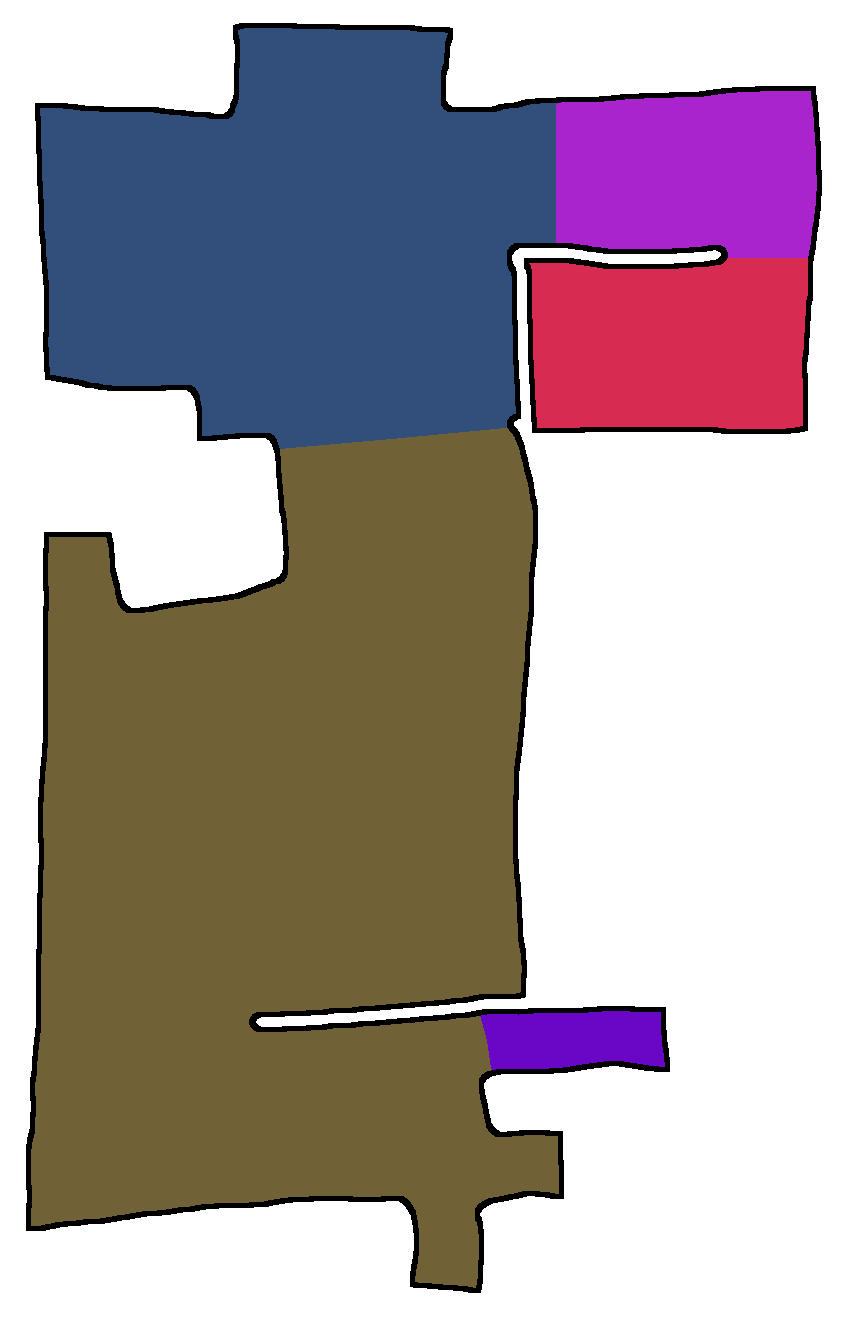}}\,
\subfloat[DuDe]{\includegraphics[width = 0.23\columnwidth]{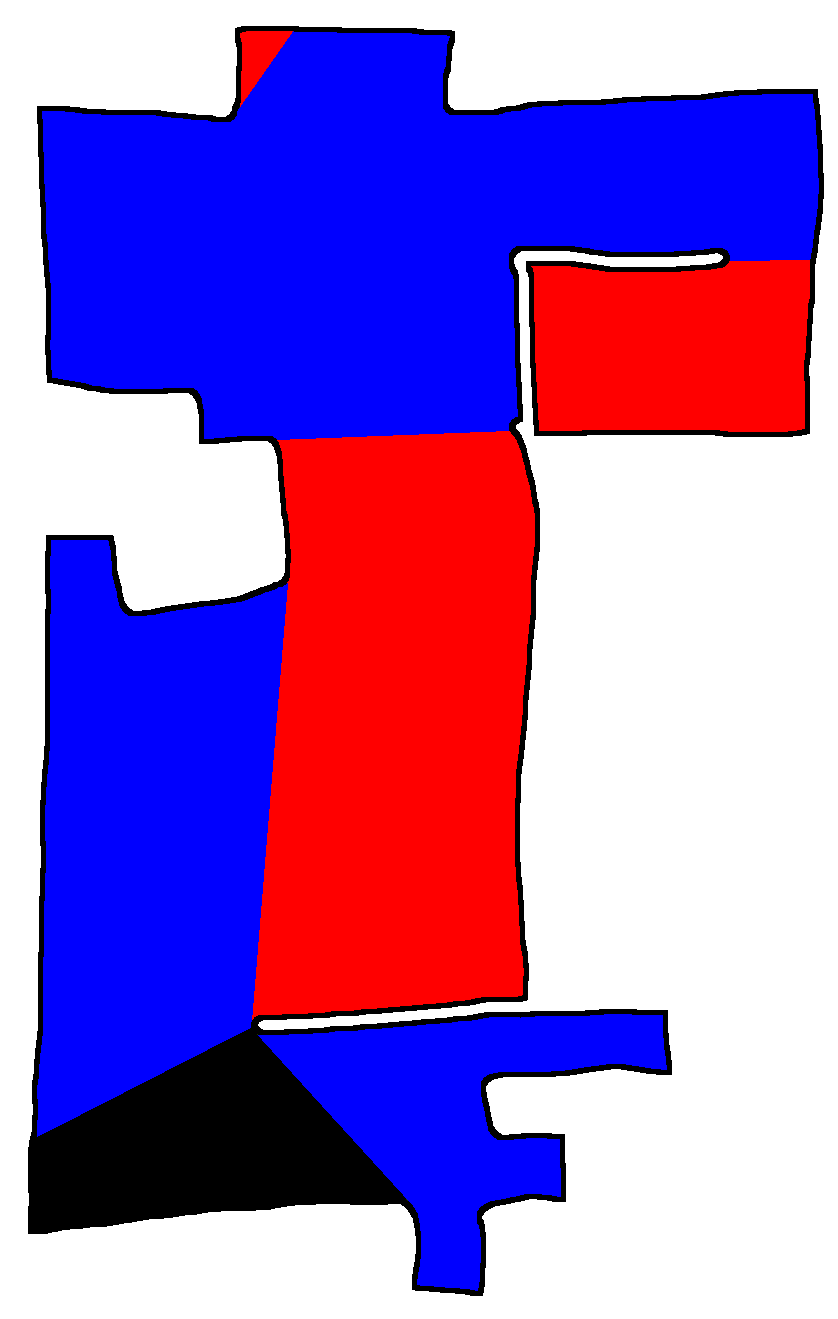}}\,
\subfloat[Ground truth]{\includegraphics[width = 0.24\columnwidth]{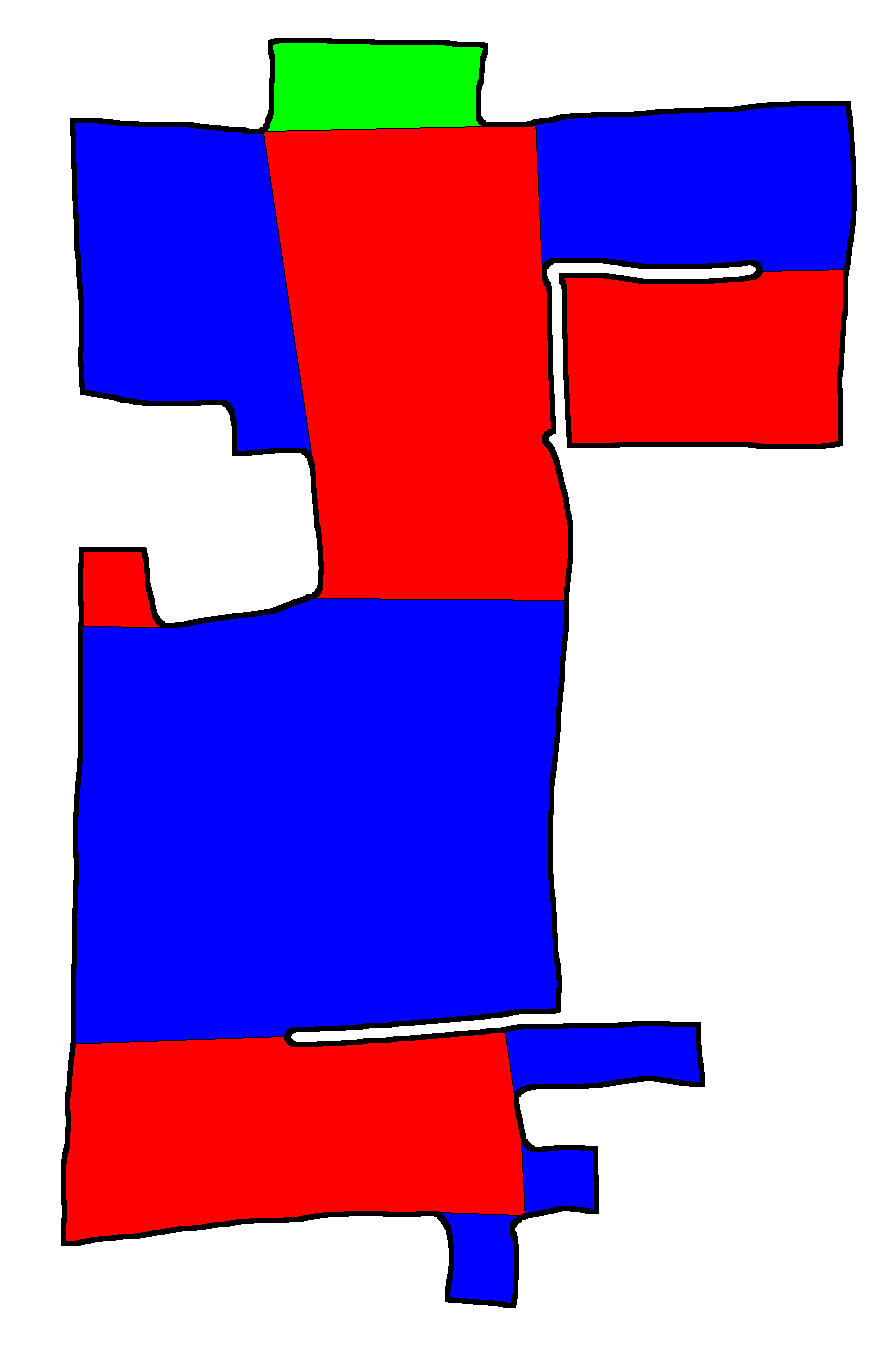}}\\
\subfloat[\textsc{maoris}]{\includegraphics[width = 0.23\columnwidth]{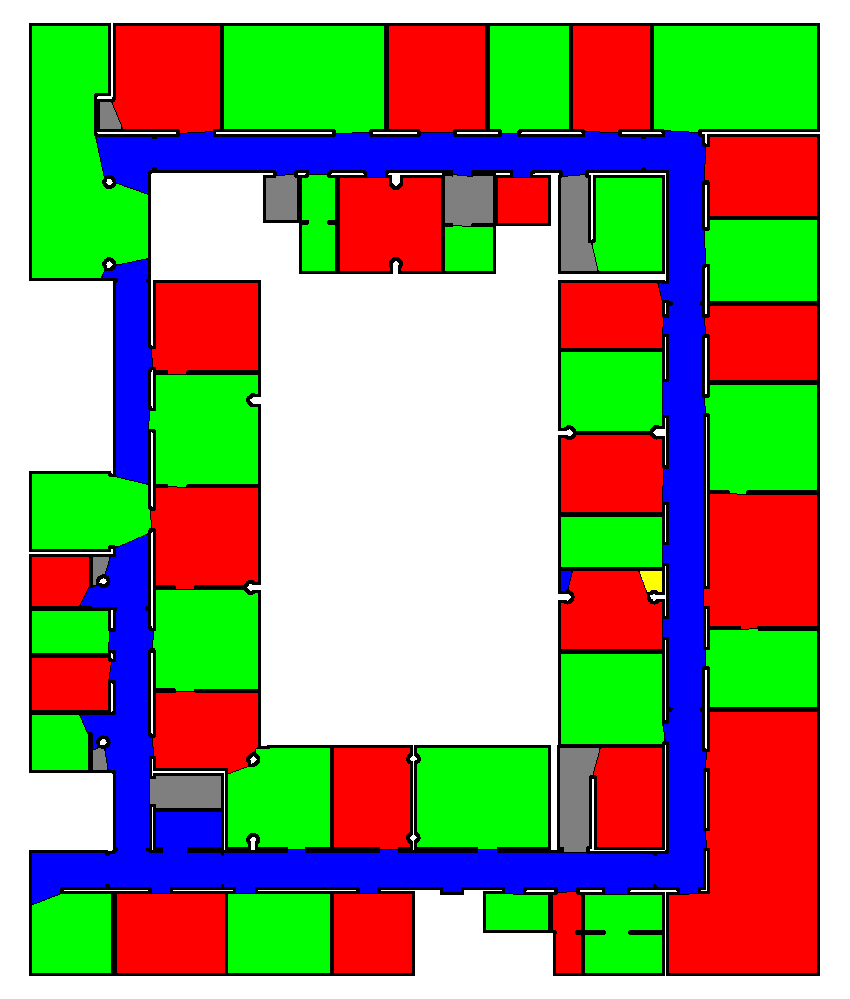}\label{fig:corridor}}\,
\subfloat[Voronoi]{\includegraphics[width = 0.23\columnwidth]{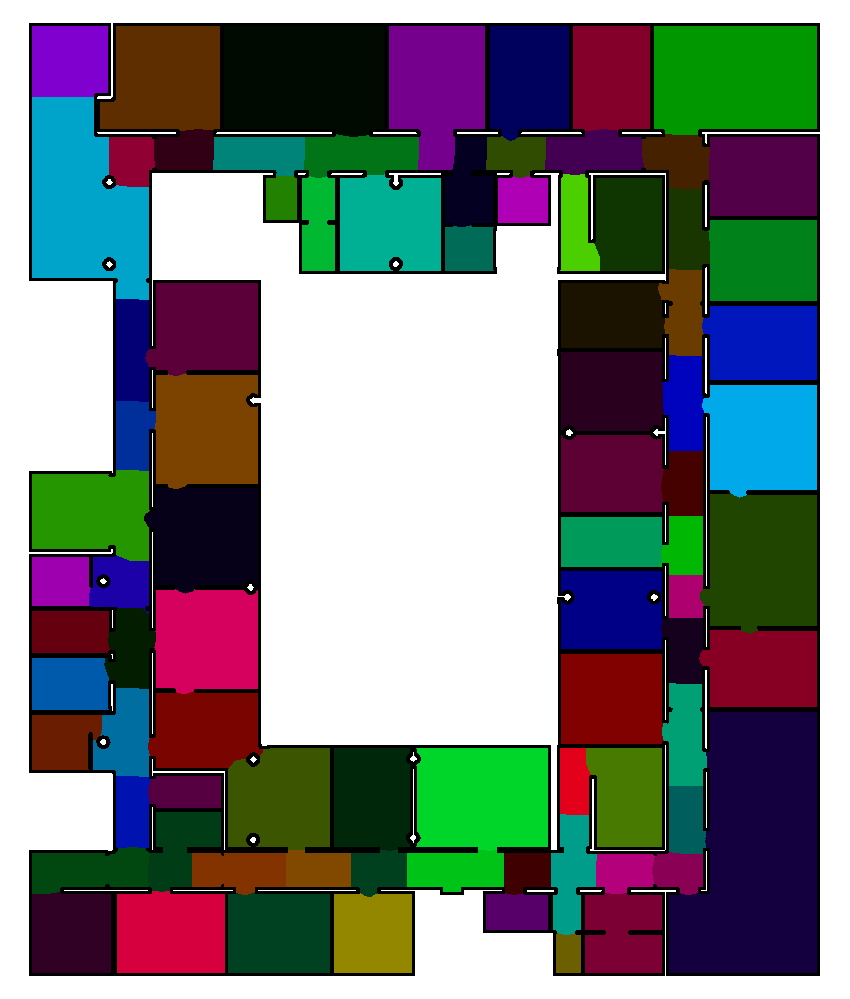}}\,
\subfloat[DuDe]{\includegraphics[width = 0.23\columnwidth]{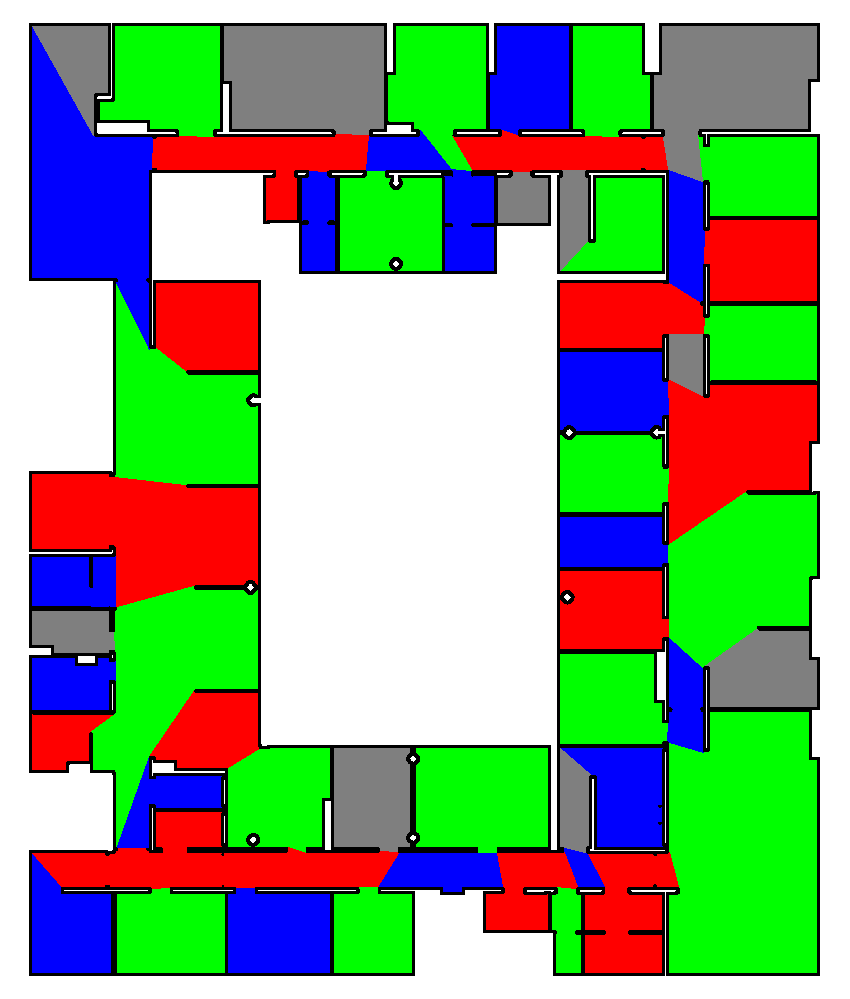}\label{fig:final}}\,
\subfloat[Ground truth]{\includegraphics[width = 0.23\columnwidth]{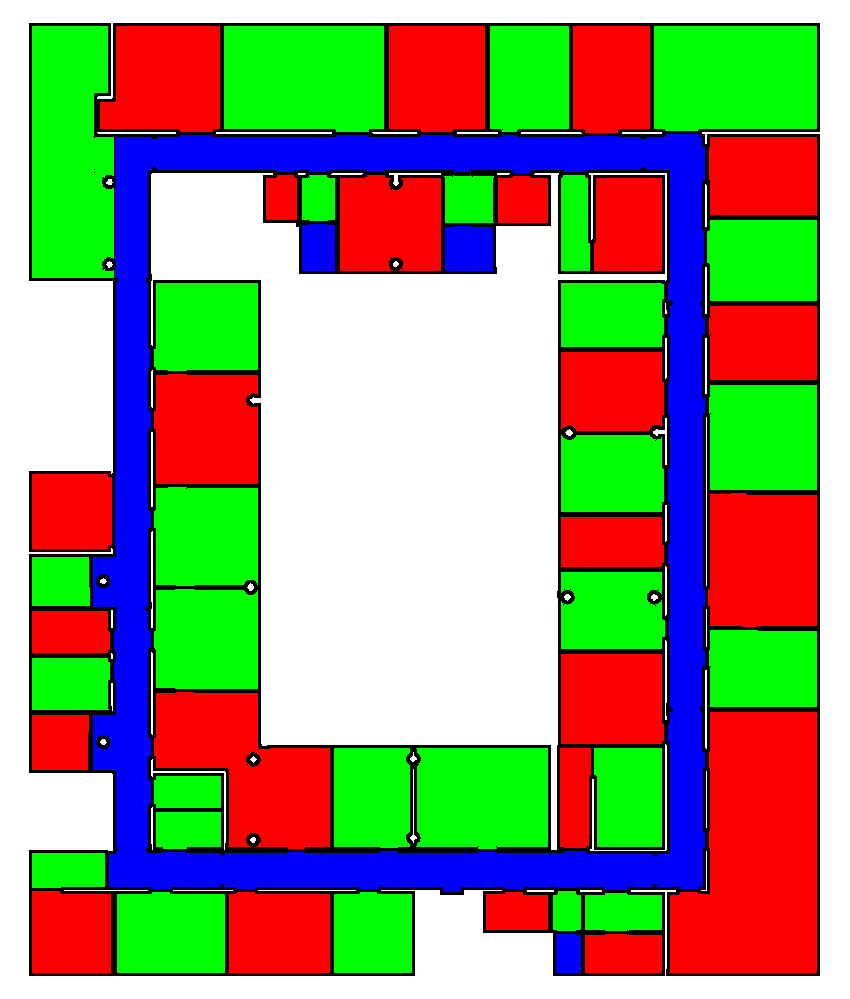}\label{fig:final}}\,
\subfloat[\textsc{maoris}]{\includegraphics[width = 0.23\columnwidth]{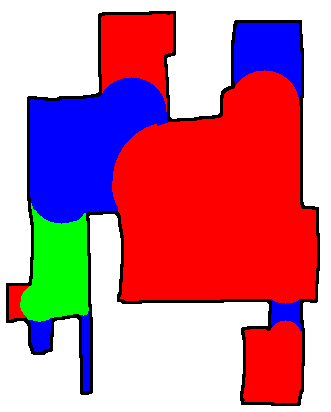}}\,
\subfloat[Voronoi]{\includegraphics[width = 0.23\columnwidth]{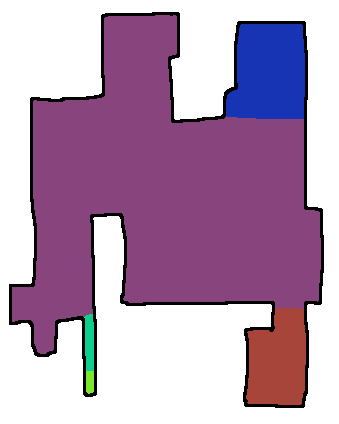}}\,
\subfloat[DuDe]{\includegraphics[width = 0.23\columnwidth]{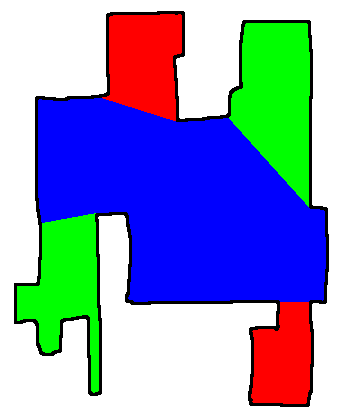}}\,
\subfloat[Ground truth]{\includegraphics[width = 0.23\columnwidth]{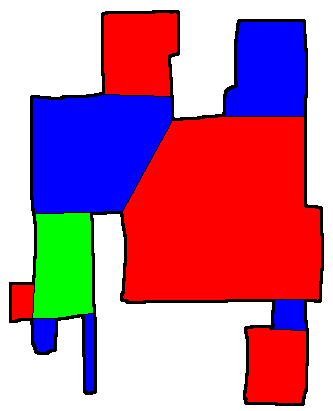}}\,
%
\caption{Examples of segmentation results for all algorithms and associated ground truths. Left: maps from Bormann's dataset. Right: maps from our dataset of sketches. One can see that \textsc{maoris} performed better than the other methods. For example the corridor in \cref{fig:corridor} is almost entirely extracted as one region while the two other methods over segment it.}
\vspace{-7mm}
\label{fig:segex}
\end{figure*}

Examples of segmentations given by all three algorithms on maps from our dataset of sketches can be seen in \cref{fig:segex}.


%


\section{Limitations and future work}

\textsc{maoris}' speed depends on the image size. Since the method is pixel based, the more pixels in the image, the slower the method. This drawback can be easily overcome by reducing the resolution of large images. Over the datasets of \textcite{bormann_room_2016}, \textsc{maoris} has a mean processing time of 43s when run on an Intel i7-4712HQ at 2.30GHz with 16GB of RAM.
However, most of the processing time comes from 2 large maps.
Without those 2 maps, the average processing time is 8.8s. Over the Bormann's dataset, with an Intel Core i7 CPU at 2.70GHz, the DuDe method processed 95\% of the maps in under 1s~\cite{fermin-leon_incremental_2017} and the Voronoi segmentation took about 13s~\cite{bormann_room_2016}.
Over the sketch dataset \textsc{maoris} has a mean processing time of 6s.
Developing an incremental method of segmentation for \textsc{maoris} would increase the processing speed. We leave this for future work.

Another way \textsc{maoris} could be improved is by being able to handle clutter in maps. In the future, we plan to work on ways to automatically remove clutter from maps before segmentation, allowing us to segment cluttered maps. We also plan to use regions given by the \textsc{maoris} segmentation method to perform matching between maps from different modalities.

\section{Summary}

We developed a strategy to segment maps from different modalities, and evaluated it on robot built and sketch maps. While robot maps are metrically accurate representations of an environment they suffer from noise in the measurements. On the other hand, sketch maps represent the topology of the environment but are not metrically accurate and contain inaccuracies, sometimes even done purposely by the user.

We also identified a flaw in the way segmentation have been evaluated in recent works and proposed a more consistent metric based on Matthew's correlation coefficient.


We evaluate our algorithm against state of the art algorithms for map segmentation, on two datasets: one standard benchmark dataset of robot maps, and one dataset of hand-drawn sketch maps that we make publicly available. Our method outperforms the state of art algorithms and obtains better Matthew's correlation coefficients for both.

\printbibliography 

\end{document}